\documentclass[a4paper,10pt,twocolumn]{article}
\usepackage[utf8]{inputenc}
\usepackage{titlesec}
\usepackage{multicol,graphicx}
\usepackage{dblfloatfix}
\usepackage{subcaption}
\usepackage{mathptmx}
\usepackage[T1]{fontenc}
\usepackage{textcomp}
\usepackage{url}
\usepackage{lipsum}

\usepackage[T1]{fontenc}
\usepackage[english]{babel}

\usepackage[backend=bibtex,style=ieee,doi=false,isbn=false,url=true,maxnames=6,citestyle=numeric-comp,giveninits=true]{biblatex}
\bibliography{references.bib}
\AtEveryBibitem{%
    \clearfield{urldate}%
    \clearfield{urlyear}%
}

\usepackage{xcolor}
\usepackage{hyperref}

\fontfamily{ptm}\selectfont

\usepackage[font=small]{caption}
\usepackage{amsmath}

\usepackage{nameref}

\makeatletter
\newcommand\setcurrentname[1]{\def\@currentlabelname{#1}}
\makeatother

\newcommand{\mysection}[1]{\vspace{0.4cm} \uppercase{#1}\setcurrentname{#1}\phantomsection \vspace{0.4cm}}
\newcommand{\mysubsection}[1]{\hspace{10pt}\textit{#1}\setcurrentname{#1}\phantomsection}

\hypersetup{
pdftitle={S-JEPA: towards seamless cross-dataset transfer through dynamic spatial attention},
pdfauthor={Pierre Guetschel},
pdfkeywords={EEG, BCI, SSL, JEPA, Masking, S-JEPA}
}

\begin{document}


\setlength{\textfloatsep}{10pt plus 1.0pt minus 2.0pt}	
\setlength{\columnsep}{1cm}


\twocolumn[%
\begin{@twocolumnfalse}
\makeatletter
\begin{titlepage}
\begin{center}
	{\fontsize{14}{18}\selectfont
        \textbf{\uppercase{S-JEPA: towards seamless cross-dataset transfer through dynamic spatial attention}}\\}
    \begin{large}
        \vspace{0.6cm}
        Pierre Guetschel\textsuperscript{1}, 
        Thomas Moreau\textsuperscript{2}, 
        Michael Tangermann\textsuperscript{1}\\
        \vspace{0.6cm}
        \textsuperscript{1}Donders Institute for Brain, Cognition and Behaviour,
        Radboud University, Nijmegen, Netherlands\\
        \textsuperscript{2}Université Paris-Saclay, Inria, CEA, Palaiseau 91120, France\\
        \vspace{0.5cm}
        E-mail: \href{mailto:pierre.guetschel@donders.ru.nl}{pierre.guetschel@donders.ru.nl}
        \vspace{0.4cm}
    \end{large}
\end{center}	
\end{titlepage}
\makeatother
\end{@twocolumnfalse}%
]%


ABSTRACT: 
Motivated by the challenge of seamless cross-dataset transfer in EEG signal processing, this article presents an exploratory study on the use of Joint Embedding Predictive Architectures (JEPAs). In recent years, self-supervised learning has emerged as a promising approach for transfer learning in various domains. However, its application to EEG signals remains largely unexplored. 
%
In this article, we introduce Signal-JEPA for representing EEG recordings which includes a novel domain-specific spatial block masking strategy and three novel architectures for downstream classification. The study is conducted on a 54~subjects dataset and the downstream performance of the models is evaluated on three different BCI paradigms: motor imagery, ERP and SSVEP. 
Our study provides preliminary evidence for the potential of JEPAs in EEG signal encoding. Notably, 
our results highlight the importance of spatial filtering for accurate downstream classification and reveal an influence of the length of the pre-training examples but not of the mask size on the downstream performance.


\begin{figure*}[!t]
    \raggedleft
    \vspace{-1ex}
    \includegraphics[scale=.9]{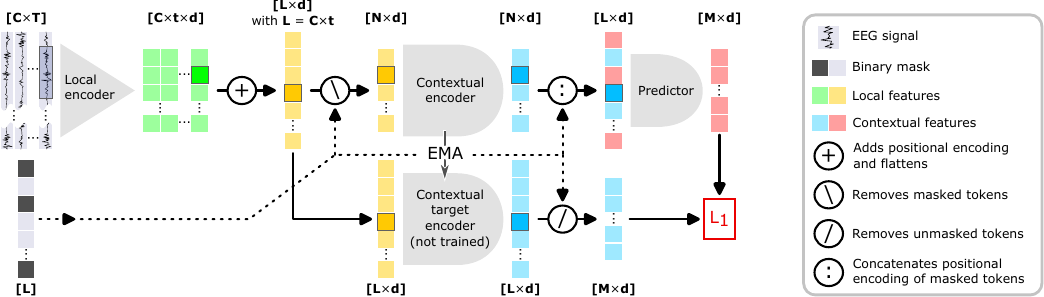}
    \caption[S-JEPA]{
    \textbf{S-JEPA} training procedure. 
    The framework takes as input EEG recordings with $C$ channels and $T$ time samples, and binary masks of length $L$. 
    First, the \textit{Local encoder}  independently transforms $t$ windows from each channel into $C\times t=L$ embedding vectors, called \textit{tokens}, of dimensionality $d$.
    Then, the tokens are marked according to their originating channel and temporal position, and are flattened into a sequence of length $L$.
    Subsequently, only the unmasked tokens are passed to the \textit{Contextual encoder}, while the full tokens sequence is given to the \textit{Contextual target encoder} to generate training targets.
    Finally, the \textit{Predictor} attempts to reconstruct the masked tokens and its predictions are compared with the targets using an L\textsubscript{1} loss.
    During the optimisation, the parameters of the Contextual target encoder 
    are not trained via gradient backpropagation but 
    follow those of the Contextual encoder by Exponential Moving Average (EMA).
    Figure inspired from~\cite{bardesRevisitingFeaturePrediction2024}.}
    \label{jepa}
\end{figure*}

\mysection{Introduction}\label{intro}

Electroencephalography (EEG) allows capturing neural activity directly from the scalp, offering a high temporal resolution signal to investigate brain functions.
The interpretation of EEG signals with machine learning methods opens the door to build brain-computer interfaces (BCIs). 
Despite the potential of BCIs, their practical application is hindered by the intensive requirement for calibration data, which is both time-consuming and demanding for participants. 
Recently, transfer learning has been explored to mitigate the constraints imposed by data scarcity and calibration demands.
Self-supervised learning (SSL) presents itself as a pivotal strategy to tackle transfer learning, enabling models to learn rich representations from unlabeled data that can be used to efficiently solve  downstream tasks.
A particular approach to SSL consists in strategically masking parts of the input data and training a model to predict these masked elements.
A key advantage of masking-based SSL methods is that they can be applied to virtually any type of data.

\mysubsection{Masking Strategies}
proved a determining factor for the success of SSL methods in various domains, including image~\cite{baevskiEfficientSelfsupervisedLearning2022,heMaskedAutoencodersAre2022,assranSelfSupervisedLearningImages2023}, speech~\cite{baevskiWav2vecFrameworkSelfSupervised2020,baevskiEfficientSelfsupervisedLearning2022}, text~\cite{baevskiEfficientSelfsupervisedLearning2022}, and video~\cite{bardesRevisitingFeaturePrediction2024} processing.
For instance, \textit{random masking} strategies, where the regions to be masked are sparsely selected, typically deliver inferior results compared to \textit{block masking} strategies, where larger continuous regions are masked, requiring the model to gain a deeper understanding of the data distribution~\cite{baevskiEfficientSelfsupervisedLearning2022}.

\mysubsection{Masked Autoencoders (MAEs)}
stand as the quintessential entry point to masking-based SSL~\cite{heMaskedAutoencodersAre2022}.
These models aim to reconstruct the masked sections of the input directly, and the training objective is computed by comparing the original input sections to the reconstructed ones.
Unfortunately, reconstructing the input and comparing elements in the original space is not without challenges.
When the original input space has a high dimensionality, the reconstruction can be computationally expensive, and necessitate the use of domain-specific constraints to produce valid signals.
The original signals can also be noisy, increasing the difficulty of encoding meaningful parts of the signal.
Moreover, the reconstruction's difficulty can vary significantly across different areas of the input data: with images, reconstructing a monochromatic, non-structured sky is less difficult than reconstructing a structured object like a hand.
Such structural disparities are one of the reasons, why conventional metrics like mean-square error often fail to assess the reconstruction quality.
These challenges question the scalability and adaptability of MAEs as a universal SSL approach.

\mysubsection{Joint-Embedding Predictive Architectures (JEPAs)}
offer a promising alternative to address the limitations associated with direct reconstruction.
JEPA-like methods avoid reconstructing the input in its original space and focus on predicting latent representations, or \textit{embeddings}, of the data~\cite{assranSelfSupervisedLearningImages2023}.
This approach confers two major benefits: first, it is computationally efficient, especially with high-dimensional input spaces as the embeddings can reduce the dimensionality; second, the metric's selection in the embedding space is less critical, as the embeddings are learned adaptively to the chosen metric.
However, as the "ideal" embedding vectors are unknown, the reconstruction objective is undefined a priori.
This challenge is addressed by constructing target embedding vectors during the training by using a bootstrapping procedure which will be further explained in the \nameref{method} section.
The potential of JEPA-like frameworks has been highlighted by their promising results with images~\cite{baevskiEfficientSelfsupervisedLearning2022,assranSelfSupervisedLearningImages2023}, speech~\cite{baevskiWav2vecFrameworkSelfSupervised2020,baevskiEfficientSelfsupervisedLearning2022}, text~\cite{baevskiEfficientSelfsupervisedLearning2022}, and videos~\cite{bardesRevisitingFeaturePrediction2024}.

\mysubsection{Applications to the EEG Domain} of
masking-based SSL techniques have started to emerge.
Pérez-Velasco and colleagues used a masked autoencoder approach, with random masking over the spatial and temporal dimensions~\cite{perezvelascoEvaluacionImpactoAprendizaje2023}.
Chien and colleagues experimented with MAEs with block masking over the temporal dimension~\cite{chienMAEEGMaskedAutoencoder2022}. 
Kostas et al.~and Foumani et al.~also used block masking over the temporal dimension but coupled it with JEPA-like training strategies~\cite{kostasBENDRUsingTransformers2021, foumaniEEG2RepEnhancingSelfsupervised2024}. 

\mysubsection{Motivation.}
Despite the advancements in applying SSL to EEG data, the exploration of block masking strategies over EEG channels remains uncharted territory. Such an approach holds the potential for developing robust channel attention mechanisms, and thus could facilitate dynamic spatial filtering. This capability could prove instrumental in adapting to recordings with varying channel sets, thereby facilitating cross-dataset transfer learning or tackling corrupted channels.
This paper seeks to bridge this gap by investigating the implications of a channel-based block masking strategy on SSL efficacy in EEG signal processing.

The main application of models trained with SSL is the following fine-tuning on the actual task of interest, the so-called downstream task. While most research within the EEG domain has focused on fine-tuning for sleep stage classification tasks, their application to a BCI context remains largely untapped, with only two studies by Kostas et al.~and Pérez-Velasco et al.~exploring the impact of SSL on BCI tasks~\cite{kostasBENDRUsingTransformers2021,perezvelascoEvaluacionImpactoAprendizaje2023}.
As BCI systems suffer from data scarcity there is a consistent goal to minimize the amount of calibration data required before each online session, specifically as the calibration phase requires sustained attention from the participant and thus is tiring. 
This highlights a significant opportunity to explore the effectiveness of SSL models across various BCI paradigms, including but not limited to motor imagery protocols, thereby contributing to the broader understanding and application of SSL in enhancing BCI performance.

Furthermore, the application of pre-trained SSL models for solving downstream tasks often involves the addition of a linear layer atop the embedding dimension. This practice, however, may not be optimal in high-dimensional embedding spaces. Through a comparative analysis of six different strategies for leveraging pre-trained architectures in downstream tasks, this work aims to support our understanding and application of SSL in EEG data processing.

\mysubsection{Research Questions and Plan.}
This manuscript is an explorative study investigating what approaches should be adopted for training SSL algorithms on EEG signals, and what domain-specific considerations warrant attention.
We approach this through three research questions:
1) What constitutes the most efficacious masking strategy for SSL when applied to EEG data?
2) How does the temporal length of examples used influence the SSL training process?   
3) What fine-tuning strategies lead to the best downstream performance?

To answer these, we first propose a novel masking strategy as part of the Signal-based Joint-Embedding Predictive Architecture (S-JEPA) framework in the \nameref{method} section. Then, we introduce fine-tuning strategies tailored for S-JEPA in the \nameref{evaluation} section. The datasets used for pre-training and for downstream evaluation are presented in the \nameref{materials} section. Finally, the \nameref{results} and \nameref{discussion} sections will report and critically discuss the outcomes and implications of the experiments we conducted.

\mysection{S-JEPA Framework}\label{method}

The Signal-based Joint-Embedding Predictive Architecture (S-JEPA) framework is illustrated in \autoref{jepa}. It is used to pre-train models. Its architecture is inspired by BENDR and MAEEG, introduced in the pioneering studies by respectively Kostas et.~al.~\cite{kostasBENDRUsingTransformers2021} and Chien et.~al.~\cite{chienMAEEGMaskedAutoencoder2022}. A key modification is introduced in the design of the local encoder to enable our novel masking strategy. This section details the architecture's components, the spatial masking strategy, and the training process.

\mysubsection{Local encoder.}
The local encoder is implemented as a convolutional neural network (CNN) with five layers, each formed of a convolution with a Gaussian error linear unit (GELU) non-linearity. Contrary to the encoders in BENDR and MAEEG, which accept multi-channel input windows, our encoder processes windows from a singular channel. 
Each window is encoded into a 64-dimensional embedding vector, hereafter referred to as a \textit{token}. 
The windows are 1.19\,s long, with a stride of 1.0\,s. 
The first convolutional kernel covers 0.25\,s while the following ones simply combine feature vectors in pairs, i.e., both the kernel temporal lengths and strides are 2. 

\mysubsection{Contextual encoder.}
This encoder consists of a transformer architecture with eight layers, as introduced by Vaswani and colleagues~\cite{vaswaniAttentionAllYou2017}. It processes the unordered sequence of tokens generated by the local encoder, necessitating the addition of position-encoding information to localize them temporally and spatially. 
The temporal positioning of each token is defined using a cosine encoding~\cite{vaswaniAttentionAllYou2017} over the first 34 dimensions, 
whereas spatial positioning is achieved through trainable embeddings for each channel, initialized using cosine encoding based on their three-dimensional coordinates. 
The contextual encoder receives tokens that only contain local information; its role is to establish relationships between them.

\newcommand{\channelSubfig}[1]{
\begin{subfigure}{0.28\linewidth}
    \centering
    \includegraphics[width=\linewidth]{illustrations/channels_masks_2d_#1.pdf}
    \caption{#1 as center.}
    \label{spat_mask_#1}
\end{subfigure}
}
\begin{figure}[!b]
    \centering
    \channelSubfig{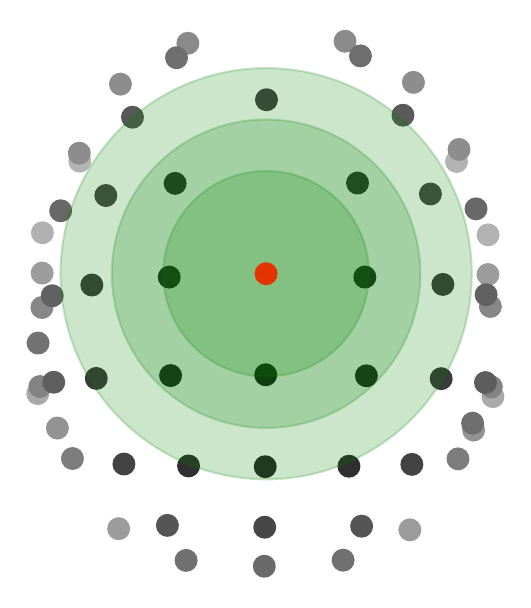}\hfill%
    \channelSubfig{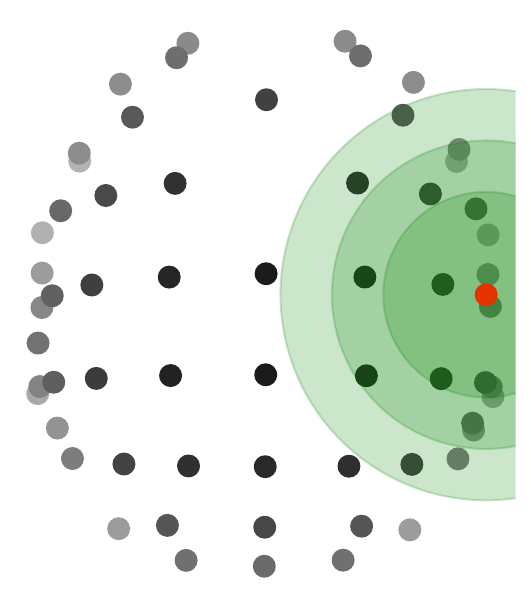}\hfill%
    \channelSubfig{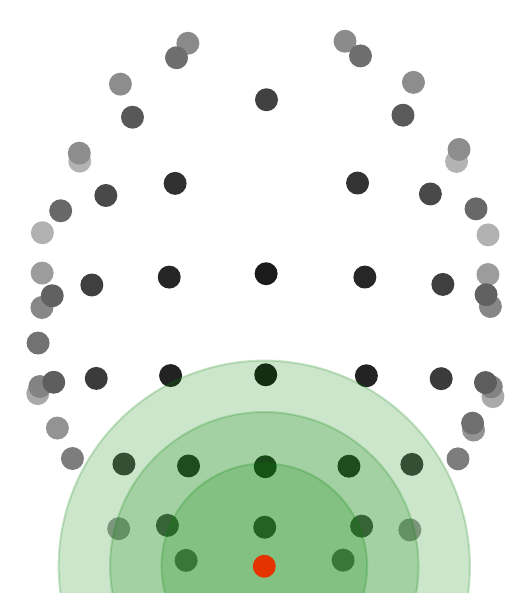}
    \caption[Spatial Block Masking]{
        Visualisation of the \textbf{spatial block masking} strategy for three example mask centres (red electrodes).
        The dark to light green spheres represent masks of diameters 40\,\%, 60\,\% and 80\,\% of the head size, as used in our experiments.
        Assuming a top-down view upon the scalp, the depth of the electrodes is denoted by their intensity (black: close, grey: distant).
        For a given mask, all electrodes within the corresponding sphere are hidden from the contextual encoder and must be predicted by the predictor.
    }
    \label{spat_mask}
\end{figure}

\mysubsection{Spatial Block Masking strategy.}
Unique to our methodology is the independent encoding of each channel by the local encoder. While it avoids the learning of spatial filters at early stages, it paves the way for innovative spatial dimension masking strategies. Literature in both, image processing~\cite{baevskiEfficientSelfsupervisedLearning2022} 
and EEG signal analysis~\cite{foumaniEEG2RepEnhancingSelfsupervised2024} suggests that block masking yields superior results compared to random masking. Motivated by these findings, our novel approach extends block masking to the spatial domain of EEG channels.

Given the irregular distribution of EEG channels, the concept of a contiguous block of tokens does not translate directly from its application in images or temporal signals, where the pixels and time samples are regularly spaced. 
Thus our approach masks all channels within a predetermined radius of a randomly chosen central channel, as illustrated in \autoref{spat_mask}. 
In this work, we compare three mask sizes with diameters approx.~40\,\%, 60\,\% and 80\,\% of
the head size.
This strategy inherently introduces variability in the number of masked tokens.



\mysubsection{S-JEPA pre-training.}
With the operational principles of both the local and contextual encoders established, we introduce two ancillary components exclusively utilized during the training phase. The first, termed the \textit{Contextual target encoder}, is a non-trainable duplicate of the contextual encoder which serves to generate the training targets. Its parameters are updated via exponential moving average (EMA). 
The second component, the \textit{Predictor}, is a transformer decoder architecture with four layers as delineated by Vaswani et al.~\cite{vaswaniAttentionAllYou2017}.
The comprehensive training methodology is illustrated in \autoref{jepa}.

%
The models are trained until no improvement of the validation loss is observed for 10 epochs (i.e., complete pass through the entire dataset), at which point the best model is saved for subsequent fine-tuning.

\mysection{Downstream Evaluation} \label{evaluation} 

Upon completion of the network's pre-training through the SSL task, which holds no intrinsic value beyond training purposes, we proceed to evaluate its efficacy on practical \textit{downstream} classification tasks which, in our case, are BCI tasks. 
This step is crucial for determining the real-world applicability of the pre-trained model.

\begin{figure}[t]
    \begin{subfigure}{\linewidth}
        \includegraphics[scale=.85]{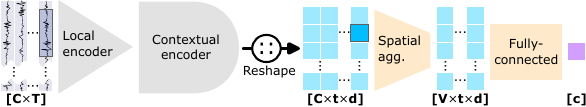}
        \caption[Contextual architecture]{\textbf{Contextual} downstream architecture.
        \label{contextual_arch}
        }
    \end{subfigure}
    \vspace{1.5ex}
    
    \begin{subfigure}{\linewidth}
        \includegraphics[scale=.85]{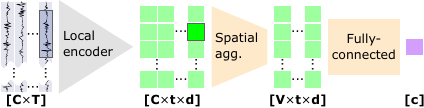}
        \caption[Post-local architecture]{\textbf{Post-local} downstream architecture.
        \label{post-local_arch}
        }
    \end{subfigure}%
    \vspace{1.5ex}

    \begin{subfigure}{\linewidth}
        \includegraphics[scale=.85]{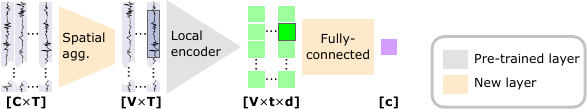}
        \caption[Pre-local architecture]{\textbf{Pre-local} downstream architecture.
        \label{pre-local_arch}
        }
    \end{subfigure}
    \caption[Downstream classification architectures]{
    \textbf{Downstream classification architectures}.
    In each of the three alternative alterations of the pre-trained networks, two new layers are added. 
    1) \textit{Spatial aggregation} is a convolutional layer that realizes weighted combinations of the elements in the channels dimension into $V\ll C$ "virtual" channels. 
    2) \textit{Fully-connected} is a linear layer that predicts $c$ class probabilities.
    }
    \label{downstream_arch}
\end{figure}

\mysubsection{Downstream classification architectures.}
Using pre-trained models for downstream classification tasks necessitates altering their architecture to allow predicting class probabilities. 
The most widely adopted modification, linear probing,  consists of adding a linear classification layer directly above the embedding space~\cite{assranSelfSupervisedLearningImages2023}. 
However, the individual tokenization of each channel in our approach leads to a high-dimensional latent space, close to the dimensionality of the raw input examples, which would make linear probing inefficient.
In response to this challenge, we enrich the architecture with two layers instead which are \textit{Spatial aggregation} and \textit{Fully-Connected}, as explained in \autoref{downstream_arch}. 

The integration of these layers is explored in three distinct configurations. 
(\subref*{contextual_arch}) The \textbf{Contextual} downstream architecture places both layers after the contextual encoder as depicted in \autoref{contextual_arch}.
(\subref*{post-local_arch}) The \textbf{Post-local} downstream architecture discards the pre-trained contextual encoder and adds the novel layers atop the local encoder as shown in \autoref{post-local_arch}.
(\subref*{pre-local_arch}) The \textbf{Pre-local} downstream architecture also discards the pre-trained contextual encoder but then places the spatial averaging layer \textit{before} the local encoder as illustrated in \autoref{pre-local_arch}. This third alternative allows the network to perform a spatial EEG filtering step, as commonly present in BCI architectures.

\pagebreak

\mysubsection{Fine-tuning.}
We examine two distinct fine-tuning strategies for these downstream architectures. 
The first method, indicated by the prefix \textit{new-}, involves exclusively training the newly introduced layers, keeping the pre-trained components frozen. 
The alternative strategy, denoted with the prefix \textit{full-}, consists in fine-tuning the entire network. This second strategy starts with a warm-up phase of 10 epochs where only the newly added layers undergo training, preventing the deterioration of the pre-trained layers' performance due to irrelevant feedback~\cite{kumarFineTuningCanDistort2022}, before the previously existing layers are included into the training.

For both strategies, the model is fine-tuned until no improvement of the validation loss is observed for 50 epochs, at which point the best model is restored for testing.
Additionally, it should be noted that the temporal length of examples used during the fine-tuning phase is determined by the requirements of the downstream task, which is independent of the length of examples utilized during the SSL pre-training phase.

\mysection{Datasets}\label{materials}

For the exploratory investigation in this work, we used the dataset introduced by Lee and colleagues~\cite{leeEEGDatasetOpenBMI2019}, subsequently referred to as the \textit{lee2019} dataset. 
It contains EEG recordings from 54 subjects, each undergoing three distinct BCI paradigms: steady-state visual evoked potentials (SSVEP) with four classes, visual event-related potentials (ERP), and left vs.~right hand motor imagery (MI). 
The recordings are from $C=62$ EEG channels, the spatial distribution of which is detailed in \autoref{spat_mask}. 
The dataset was loaded from the MOABB framework~\cite{aristimunhaMotherAllBCI2023}, bandpass filtered at 0.5\,-\,40\,Hz and downsampled to~128\,Hz.

We used the first 40 subjects to pre-train any model. The subsequent 7 subjects were used for the validation during this pre-training phase. The remaining 7 subjects were reserved for the downstream performance evaluation.

\mysubsection{Pre-training data.}
SSL methods do not necessitate labels. 
As such, the training and validation examples are slices of the continuous recordings taken at a fixed interval of 16.9 seconds.
This study compared three example lengths $T$ approximately distributed on a logarithmic scale, namely 1, 4, and 16\,seconds. 
All allocated subjects and all paradigms were used collectively during the pre-training phase which yielded a total of 36,576 training examples and 6,528 validation examples.
No artefact rejection method was applied.

\mysubsection{Downstream evaluation data.}
For each subject allocated for downstream evaluation and each paradigm, the fine-tuning performance of the different pre-trained models is assessed using a 5-folds within-subject stratified cross-validation procedure.
The examples used are 4.19\,seconds long for MI and SSVEP, and 1.19\,seconds long for ERP.
These lengths are chosen to respect the durations defined in the original dataset~\cite{leeEEGDatasetOpenBMI2019} while maintaining compatibility with our tokenization process.

\mysection{Results}\label{results}

\mysubsection{Experimental details.}
For pre-training, our setup compares all combinations of signal durations (1s, 4s, and 16s) and mask sizes (40\,\%, 60\,\%, and 80\,\% of head size),
along with a \textit{no-pre-training} baseline. 

For downstream performance evaluation, we assess each pre-trained model using the three downstream architectures---\textit{contextual}, \textit{post-local}, and \textit{pre-local}---and the two fine-tuning approaches---\textit{full} and \textit{new}. 
The \textit{no-pre-training} is only assessed with the \textit{full} fine-tuning approach.
A given pairing of pre-training and fine-tuning configurations is referred to as a \textit{pipeline}.

\begin{figure}[!b]
    \centering
    \includegraphics[width=\linewidth]{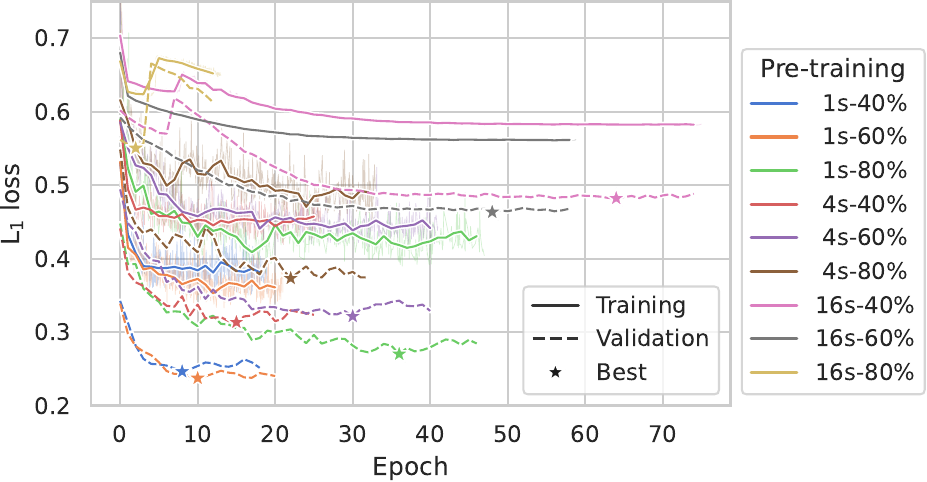}
    \caption[Training curves]{
    \textbf{Pre-training curves} of the different configurations tested. 
    The solid and dashed lines indicate the loss on the training and validation sets.
    While the validation loss was tested once per epoch only, the training loss was logged after every optimisation step. 
    The train loss on individual optimisation steps is visible in the background, corresponding epoch-wise averages are outlined in white.
    A star marks the lowest validation loss per curve, the early stopping time point and consequently the checkpoint from which any fine-tuning started.
    }
    \label{loss_plot}
\end{figure}

\pagebreak

\mysubsection{Pre-training dynamics.}
The pre-training phase, see \autoref{loss_plot}, provides insights into the training process under the S-JEPA framework. 
It reveals that training curves under the 16s condition are significantly smoother than the other configurations. 
Additionally, the early stopping mechanism concludes the training of the 16s-80\% configuration prematurely at 12 epochs due to an early trough in the loss curve. Similarly, the 16s-40\% setup also encounters an early trough yet manages to recover and complete a longer training.
The longest training durations are observed in the 16s-40\% and 16s-60\% configurations, enduring for 74 and 58 epochs, respectively, translating to approximately 12 and 10 hours of training.

\begin{figure}[!b]
    \includegraphics[width=\linewidth]{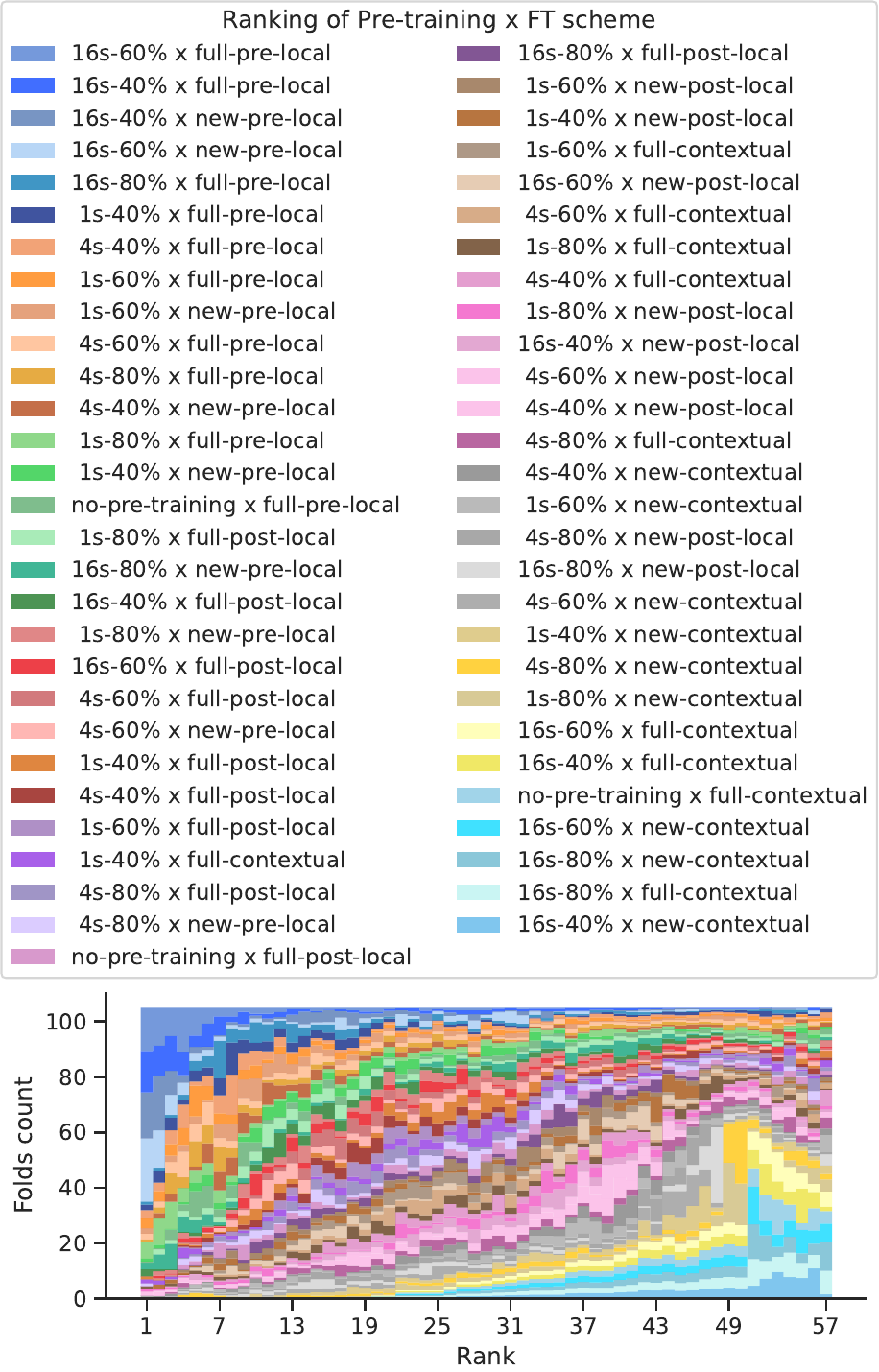}
    \caption[Ranking]{
        Global \textbf{downstream classification ranking} of all the combinations of pre-training configurations and fine-tuning schemes.
        Each of the three test datasets has 7 subjects and 5 folds per subject, which makes a total of 105 folds.
        In the legend, the combinations are ordered according to their average rank over all folds.
        The vertical span of a coloured "pixel" in the plot represents the number of folds in which this configuration has obtained the rank indicated by the x-axis.
        }
    \label{ranking}
\end{figure}

\mysubsection{Pipelines ranking on downstream performances.}
The comprehensive ranking of all tested pipelines is detailed in \autoref{ranking}, offering a comparative overview of the performance of the different combinations over the experimental protocols. 
Notably, the top-performing pipelines in the downstream tasks are the 16s-60\% and 16s-40\% models, especially when paired with any pre-local fine-tuning strategy. They notably occupy the first rank in two-thirds of the cases. 
In contrast, lower-performing pipelines feature new-contextual or full-contextual fine-tuning, particularly when combined with 16s and no-pre-training configurations.

\begin{figure}[!h]
    \centering
    \includegraphics[width=\linewidth]{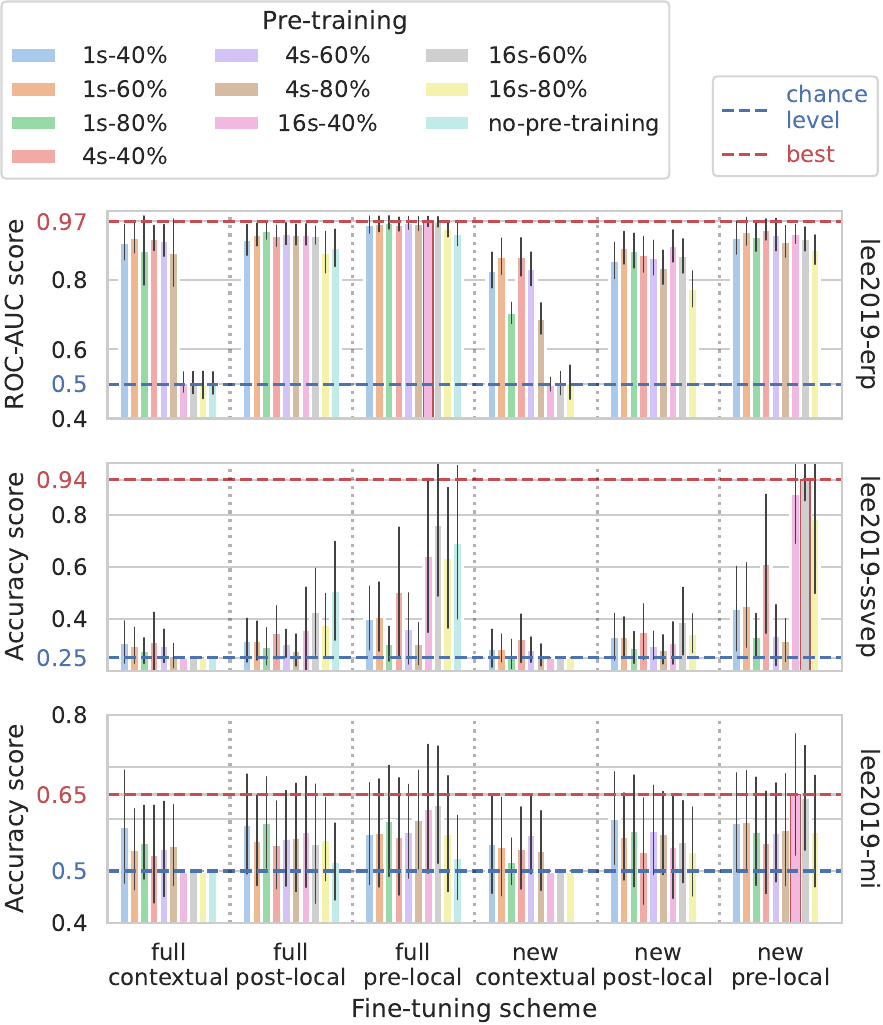}
    \caption[Downstream classification scores]{
        \textbf{Downstream classification scores} of all the pipelines on the three test datasets.
        The height of the coloured bars corresponds to the average classification score over all the test subjects and cross-validation folds, while the thin black bars correspond to their standard deviation.
    }
    \label{barplot_scores}
\end{figure}

\mysubsection{Paradigm-wise downstream performances.}
on the individual paradigms is reported in \autoref{barplot_scores}. 
For the subsequent analysis, we discern several key observations: 
1) The pipelines obtaining the best score on the ERP, SSVEP and MI tasks are respectively 16s-40\%-full-pre-local with a 97\% AUC, 16s-60\%-new-pre-local with a 94\% accuracy, and 16s-40\%-new-pre-local with a 65\% accuracy. 
2) Pipelines combining 16s or no-pre-training with contextual architectures frequently result at chance level on average.
3) Pipelines combining 1s or 4s with contextual architectures also perform at chance level on the SSVEP paradigm but above chance level on the MI and ERP ones. 
4) Most pipelines manage to achieve respectable scores on the ERP dataset. 
5) Only a select subset of pipelines excel on the SSVEP dataset. 
6) The MI paradigm scores exhibit notable variability, as indicated by the considerable standard deviation.

\mysection{Discussion}\label{discussion}


In the light of observations made in the \nameref{results} section, we aim to answer the research questions posed, draw conclusions, and provide guidelines for future research.

\mysubsection{Examples' length largely influences downstream performance.}
%
In particular, the 16s pipelines consistently show the best performance, highlighting the advantage of a longer context during pre-training.
Conversely, when considering solely the results using the contextual downstream architecture, the 16s pipelines often perform at chance level. 
On the other hand, the 1s and 4s pipelines yield better results, which may be attributed to their better alignment of the attention mechanism's training with the short signal windows of the downstream tasks.

\mysubsection{Mask radius' impact on downstream performance uncertain},
as our results do not demonstrate a clear trend between the masks compared. It is possible that the range of masks we compared is not optimal or that other factors are influencing the results. Future work should consider comparing our spatial masking strategy with temporal masking to better understand the relative strengths and weaknesses of these approaches.

\mysubsection{The best fine-tuning strategy implements spatial filtering} 
The pre-local architecture emerges as the best for downstream classification. This architecture's approach to linearly combine channels before computing features enables effective spatial filtering, thereby enhancing the signal-to-noise ratio. This finding underscores the critical role of spatial filtering in boosting model performance by leveraging the inherent spatial properties of EEG data.


\mysubsection{State-of-the-art comparison.}
According to Chevallier and colleagues~\cite{chevallierLargestEEGbasedBCI2024} who benchmarked numerous decoding algorithms across all the datasets available in MOABB~\cite{aristimunhaMotherAllBCI2023}, the current state-of-the-art (SOTA) performances for within-session classification on the Lee2019 dataset are: ERP at $98.41\pm2.03\%$, SSVEP at $89.44\pm13.84\%$, and MI at $84.74\pm13.19\%$. Notably, all pipelines establishing the SOTA utilize Riemannian geometry. Our approach matches the SOTA for ERP, enhances it for SSVEP, but falls short on MI. 
A critical difference in our evaluation methodologies should be noted: 
our downstream evaluation only focuses on the last 7 subjects, unlike their analysis on all 54 subjects. Specifically, Lee and colleagues have identified 6 out of these 7 subjects as hard to classify on the MI task~\cite{leeEEGDatasetOpenBMI2019}. We believe this exceptionally high rate of challenging subjects might explain our low MI performance.

\mysubsection{Choice of Dataset.}
The need for large datasets is paramount when training transformers, potentially explaining the underperformance of the contextual downstream strategy.
Although this exploratory study on the Lee2019 dataset provides valuable insights, future research should pivot towards larger datasets to fully harness the capabilities of contextual architectures.



\mysubsection{Conclusion.}
This exploratory work introduces a novel masking strategy and three fine-tuning approaches, positioning our method competitively within the realm of BCI tasks. We achieve SOTA performance on two out of three evaluated downstream tasks. Our findings suggest that long pre-training windows favor the local features encoder, while short windows benefit the contextual encoder. Therefore, future research should aim at successfully training both the local and contextual encoders.
However, no influence of the mask radius on the downstream performance was found.
Finally, the best downstream architecture includes a spatial filtering step and discards the contextual encoder.

\begin{small} 

\mysection{Acknowledgements}
\vspace{-1.2em}

This work 
is in part supported by the Donders Center for Cognition (DCC) and 
is part of the project Dutch Brain Interface Initiative
(DBI2) with project number 024.005.022 of the research programme Gravitation
which is (partly) financed by the Dutch Research Council (NWO).

\mysection{references}%
\vspace{-1.7em}

\renewcommand*{\bibfont}{\normalfont\small}
\printbibliography[heading=none]

\end{small}

\end{document}